\documentclass[10pt,twocolumn,letterpaper]{article}

\usepackage{iccv}              
\usepackage{float}
\usepackage{multirow}
\usepackage{array}
\usepackage{float}
\usepackage{algorithm}
\usepackage{algorithmic}
\usepackage{multirow}
\usepackage{appendix}
\definecolor{iccvblue}{rgb}{0.21,0.49,0.74}
\usepackage[pagebackref,breaklinks,colorlinks,allcolors=iccvblue]{hyperref}

\def\confName{ICCV}
\def\confYear{2025}

\title{LES-Talker: Fine-Grained Emotion Editing \\for Talking Head Generation in Linear Emotion Space}

\author{
    Guanwen Feng$^{1,2,3}$\textsuperscript{†}, Zhihao Qian$^{1,2,3}$\textsuperscript{†}, Yunan Li$^{1,2,3}$\textsuperscript{*}, Siyu Jin$^{1,2,3}$, Qiguang Miao$^{1,2,3}$\textsuperscript{*}, Chi-Man Pun$^{4}$\\
    $^{1}${\small School of Computer Science and Technology, Xidian University, Xi'an 710071, China} \\
    $^{2}${\small Xi'an Key Laboratory of Big Data and Intelligent Vision, Xi'an, Shaanxi 710071, China} \\
    $^{3}${\small Key Laboratory of Collaborative Intelligence Systems, Ministry of Education, Xidian University, Xi’an 710071, China} \\
    $^{4}${\small Department of Computer and Information Science, University of Macau, Macao 999078, China} \\[1ex]
    {\tt\small\{gwfeng\_1, zhqian\_1, syjin\}@stu.xidian.edu.cn, \{yunanli, qgmiao\}@xidian.edu.cn, cmpun@umac.mo}
}

\begin{document}

\twocolumn[{%
\maketitle
\vspace{-1.1cm}
\begin{figure}[H]
\hsize=\textwidth 
\centering
\includegraphics[width=0.95\textwidth]{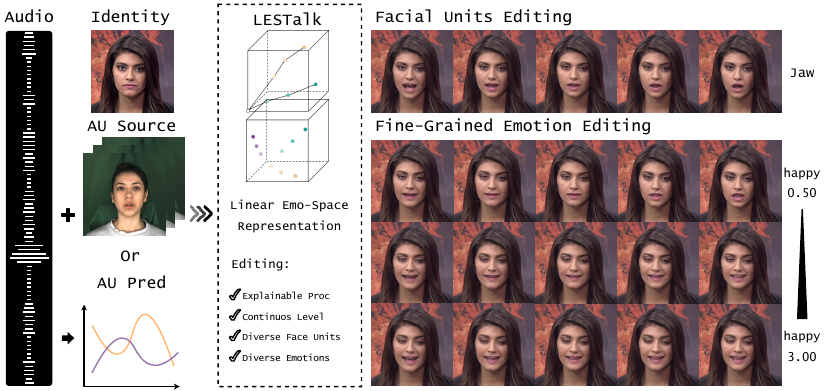}
\caption{Linear Emotion Space (LES) based on Facial Action Units (AUs) supports our LES-Talker model, offering exceptional interpretability. It enables fine-grained editing across 8 emotion types, 17 facial units, and continuous levels above 0. It can be driven by a lightweight use of video (requiring a sequence of images to provide the AU source) or by audio alone.}
\label{fig1}
\end{figure}
}]

\begin{abstract}
While existing one-shot talking head generation models have achieved progress in coarse-grained emotion editing, there is still a lack of fine-grained emotion editing models with high interpretability. We argue that for an approach to be considered fine-grained, it needs to provide clear definitions and sufficiently detailed differentiation. We present LES-Talker, a novel one-shot talking head generation model with high interpretability, to achieve fine-grained emotion editing across emotion types, emotion levels, and facial units. We propose a Linear Emotion Space (LES) definition based on Facial Action Units to characterize emotion transformations as vector transformations. We design the Cross-Dimension Attention Net (CDAN) to deeply mine the correlation between LES representation and 3D model representation. Through mining multiple relationships across different feature and structure dimensions, we enable LES representation to guide the controllable deformation of 3D model. In order to adapt the multimodal data with deviations to the LES and enhance visual quality, we utilize specialized network design and training strategies. Experiments show that our method provides high visual quality along with multilevel and interpretable fine-grained emotion editing, outperforming mainstream methods.
\href{https://peterfanfan.github.io/LES-Talker/}{Project page: https://peterfanfan.github.io/LES-Talker/}

\end{abstract}

%

\section{Introduction}
Talking head generation has attracted significant attention from researchers in recent years and has many applications in the fields of digital human creation \cite{zhang2024reduced}, virtual reality \cite{chen20223d}, etc. To enhance the diversity and expressiveness of talking head generation, emotion editing tasks were emphasized. However, existing studies have the following two notable drawbacks: (1) the interpretability of methods is lacking; (2) the effectiveness of emotion editing is still remains at a coarse-grained level.
On the one hand, some studies offer expression reference images without clear emotion definition \cite{tan2024edtalk,wang2023progressive,ji2022eamm}, while others employ discrete emotion labels \cite{gan2023efficient,tan2023emmn,ma2023styletalk}. Additionally, some extract emotion features from latent spaces \cite{wang2023progressive,peng2023emotalk,ki2024float} leading to an implicit transformation of emotion.  
Although Facial Action Units (AUs) effectively describe emotions, various methods \cite{ekman1978facial,friesen1983emfacs,ekman1998facial} use different combinations of AUs for the same emotion. Moreover, AUs alone remain insufficient. On the other hand, several studies \cite{wang2020mead,peng2023emotalk,tan2024edtalk} focus on generating videos with specific emotions, realizing only coarse-grained emotion editing. One of the few fine-grained emotion studies \cite{wang2023progressive,sun2024fg} also shows only discrete emotion levels in three emotion types.

We argue that for an approach to be considered fine-grained, it needs to offer clear definitions and detailed differentiation. Motivated by the above, we propose Linear Emotion Space (LES), a definition that explicitly characterizes emotion transformations. LES supports us in proposing LES-Talker, a video generation model for editing multiple emotion types, continuous emotion levels, and individual Facial Action Units. To provide an interpretable theoretical foundation, we develop an LES definition based on elements derived from AUs. Using a coarse-to-fine emotion strategy, LES characterizes emotion transformations as vector changes in two subspaces: the Action Subspace, representing facial unit actions, and the Isolation Subspace, capturing subtle emotion details. When fine-grained emotion editing is required, the vectors in the LES can be used as intermediate representations of emotions. Specifically, when editing a neutral emotion video, each frame is extracted with AUs values and mapped into the Action and Isolation Subspaces as vector representations. The target emotion with specified level corresponds to a vector in these spaces, allowing fine-grained emotion transformations through vector transformations. Further, each dimension of LES has a physical counterpart, making the meaning of these edits explicit.

To achieve a satisfactory editing effect, our LES-Talker utilizes LES vectors and 3DMM coefficients as two intermediate representations. We divide the task into three major components. First, we aim to adapt the multi-modal data to the LES definition. We transform the AUs sequences, enhance audios and optimize predicted 3D coefficients, through direct adaptation, wav2vec-based \cite{baevski2020wav2vec} audio encoder, and residual learning, respectively. After generating the accurate representations in LES, we apply vector operation to inject emotion according to the definition in LES. We also propose a Cross-Dimension Attention Net (CDAN). By combining CDANs in series and parallel, we enable detailed guidance of the LES representation in the deformation of the 3D model. Inspired by Sadtalker \cite{zhang2023sadtalker}, we use the identity and pose information from reference images and edited 3D coefficients to generate a video through a novel render.

Our study makes the following four main contributions:
\begin{itemize}

\item We propose the Linear Emotion Space (LES), a fine-grained emotion definition that provides an interpretable theoretical foundation for fine-grained emotion research.
\item We introduce LES-Talker, a novel one-shot talking head generation model with high interpretability, designed to achieve fine-grained emotion editing across emotion types, levels, and facial units.
\item We propose a novel and universal Cross-Dimension Attention Network to mine potential correlations between 3D model representation and LES representation, which enables the detailed guidance of LES representation in the deformation of the 3D model. 
%

\item  Quantitative and qualitative experiments demonstrate our method generates videos with high visual quality and is capable of fine-grained editing at multiple levels.
\end{itemize}

\begin{figure*}[t]
\centering
\includegraphics[width=\textwidth]{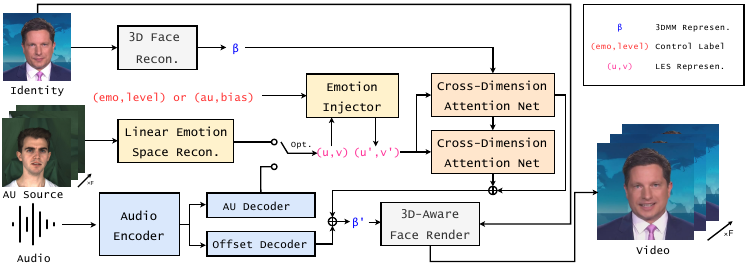} 
\caption{Pipeline of LES-Talker. Inputs include an identity image, audio, optional AU Source, and user editing targets. The Linear Emotion Space Recon. generates emotion vectors in LES. Emo Injector transforms these vectors based on user targets $(emo, level) \text{ or } (au, bias)$. Two levels of Cross-Dimension Attention Net (CDAN) process decomposed vectors $\boldsymbol{u}$ and $\boldsymbol{v}$ to produce 3D coefficients, optimized via Offset Decoder. These coefficients, along with identity information, create the rendered video.}
\label{fig2}
\end{figure*}

\section{Related work}
\subsection{Talking Head Video Generation}

Existing research on talking head generation is mainly categorized into audio-driven \cite{wang2022one,yu2023talking,sun2024diffposetalk,aneja2024facetalk,hogue2024diffted, chen2024diffsheg} and video-driven \cite{liu2021li,wang2021one,drobyshev2022megaportraits,zhao2022thin} two broad categories. In audio-driven methods, lightweight use of video to compensate the lack of information is a common approach, such as StyleTalk \cite{ma2023styletalk} extracting facial motion patterns using a transformer, and PC-AVS \cite{zhou2021pose} incorporating head pose information through implicit low-dimensional pose coding. Video-driven methods utilize the richer information contained in the input video to generate more natural results, focusing on different driving methods and intermediate representations, which are divided into implicit and explicit. Implicit modeling represents scenes by learning mathematical functions, such as Signed Distance Functions (SDF) \cite{xu2023omniavatar,zheng2023locally,shim2023diffusion} and Neural Radiation Fields (NeRF) \cite{guo2021ad,li2023efficient,shen2023sd}. Explicit modeling directly constructs editable 3D geometric representations, such as 3D Morphable Model (3DMM) \cite{zhang2023sadtalker,shen2024talking,xu2024facial} and 3D Gaussian Splatting (3D-GS) \cite{yu2024gaussiantalker,cho2024gaussiantalker,li2025talkinggaussian,cha2025emotalkinggaussian}.

We adopt LES vectors and 3DMM coefficients as representations for explicit control methods.
\subsection{Fine-grained Facial Emotion Video Generation} 
In talking head generation, early studies \cite{prajwal2020lip,cheng2022videoretalking} focused on improving visual quality, while recent studies \cite{zhang2023dream,wang2024emotivetalk,liu2025moee} have explored coarse-grained facial emotion editing. EmoTalk \cite{peng2023emotalk} employs an audio-driven method using two different audio extractors to separate emotion and content, and EDTalk \cite{tan2024edtalk} emphasizes controlled talking head generation by decoupling mouth shape, head pose, and emotional expression. However, these studies only achieve coarse-grained emotion expression.Recently, a few fine-grained emotion studies have emerged. Under co-driven conditions, PD-FGC \cite{wang2023progressive} demonstrates linear interpolation between expression features from different sources by learning the expression latent space. EmoSpeaker \cite{feng2024emospeaker} decouples audio input into content and emotion vectors, editing content vectors for different emotions. FG-EmoTalk \cite{sun2024fg} shows video-driven realization of three defined emotions and several levels of fine-grained emotions using AUs as control inputs. However, these studies lack clarity on the principles of fine-grained emotion changes and primarily realize discrete rather than continuous changes.

While utilizing AUs for fine-grained emotions is promising, it is insufficient. Our study refines elements derived from AUs, providing a rigorous definition and demonstrating talking head generation with continuously varying level and multiple emotion options, allowing individual control of specific facial units.

\section{Method} We present the Linear Emotion Space (LES) and LES-Talker that aim to achieve fine-grained emotion editing with exceptional interpretability. The complete pipeline of our one-shot method is depicted in Fig.~\ref{fig2}. Below, we first provide a brief introduction to Action Units (AUs) and 3D face models as preliminaries in Sec.~\ref{sec:preliminary}. Then, in Sec.~\ref{sec:linear emotion space}, we offer a precise and rigorous definition of the Linear Emotion Space, which serves as the theoretical foundation of our model. Cross-Dimension Attention Network (CDAN) plays a key role in enabling the LES representation to guide the controllable deformation of the 3D model, and we detail its design in Sec.~\ref{sec:cdan}. Finally, in Sec.~\ref{sec:other key components}, we describe the design of other key components that adapt multi-modal features to the LES and improve visual quality.
\setlength{\tabcolsep}{1.8mm}
\subsection{Preliminary}
\label{sec:preliminary}
\textbf{Facial Action Units.} 
Facial Action Units (AUs) are essential in studying facial expressions. AUs describe patterns of facial muscle movements associated with emotions. Currently, there are seventeen controllable AUs. For instance, in happiness, relevant AUs include AU6 (cheek muscle contraction, forming a smile), AU12 (lip corner elevation, indicating pleasure), AU25 (mouth opening, signifying excitement), and AU4 (brow raising, expressing joy or surprise). Different AUs can be used individually or in combination to represent emotional states; however, varying definitions across studies can cause confusion. Our method utilizes the complete set of AUs, along with subtle information beyond AUs, to represent each emotional state. Further explanations of AUs are in the supplementary materials.
\\\textbf{3D Morphable Model.} We use 3D Morphable Models (3DMMs) as one of our intermediate representations, inspired by the single image deep 3D reconstruction method \cite{deng2019accurate} and recent talking head generation method \cite{zhang2023sadtalker}. The 3D face shape S and the talking head motion M can be decoupled as:
\begin{equation}
\begin{array}{c}
S = \overline{S} + \alpha \cdot U_{\text{id}} + \beta \cdot U_{\text{exp}}\\
M = [\beta, \gamma]
\end{array}
\label{Eq1}
\end{equation}
$\overline{S}$ is the average shape of the 3D face. $U_{\text{id}}$ and $U_{\text{exp}}$ are the orthonormal basis \cite{10.1145/311535.311556}. Coefficients $\alpha \in \mathbb{R}^{80}$, $\beta \in \mathbb{R}^{64}$ and $\gamma \in \mathbb{R}^{6}$ describe the person identity, face expression and head pose respectively.

\subsection{Linear Emotion Space}
\label{sec:linear emotion space}
As mentioned in the Preliminary section, there are 17 controllable AUs, indexed from 1 to 17.
Their values could be extracted from the emotional frame. Through two different optimization functions (\(\text{Opt}_{\text{1}}\) and \(\text{Opt}_{\text{2}}\)), we derive a point \(\boldsymbol{w}\) with 41 coordinates, each having a specific physical meaning. In the following descriptions, points, and vectors in the LES are equivalent.%

The Linear Emotion Space is based on two hypotheses: (1) The space containing \(\boldsymbol{w}\) is linear. (2) The level of emotion \(int_{emo}\) varies linearly in this space. Therefore, the Linear Emotion Space \(\mathbb{E}\) is defined with the following characteristics:
\begin{equation}
if \ \boldsymbol{w} = (e_1, e_2, \ldots, e_{41}) \in \mathbb{E}\quad then\ \boldsymbol{w} \in \mathbb{R}^{41}
\label{Eq2}
\end{equation}
\begin{equation}
e_i = \text{Opt}_{\text{1}}(AU_i) \quad i \in [1, 17]
\label{Eq3}
\end{equation}
\begin{equation}
e_i = \text{Opt}_{\text{2}}(AU_j)\quad j=i-17\quad i \in [18, 34]
\label{Eq4}
\end{equation}
\begin{equation}
e_i = 0 \text{ except } e_{index}=od\quad i \in [35, 41]
\label{Eq5}
\end{equation}
\begin{equation}
\mathbb{A}=\{\boldsymbol{u} = (e_1,\ldots, e_{17})\} \subseteq \mathbb{E} 
\label{Eq6}
\end{equation}
\begin{equation}
\mathbb{I}=\{\boldsymbol{v} = (e_{18}, \ldots, e_{41})\} \subseteq \mathbb{E}
\label{Eq7}
\end{equation}
\begin{equation}
level_{emo} \ \text{varies linearly in}\ \mathbb{E}   
\label{Eq8}
\end{equation}
\begin{figure}[h]
\centering
\includegraphics[width=0.38\textwidth]{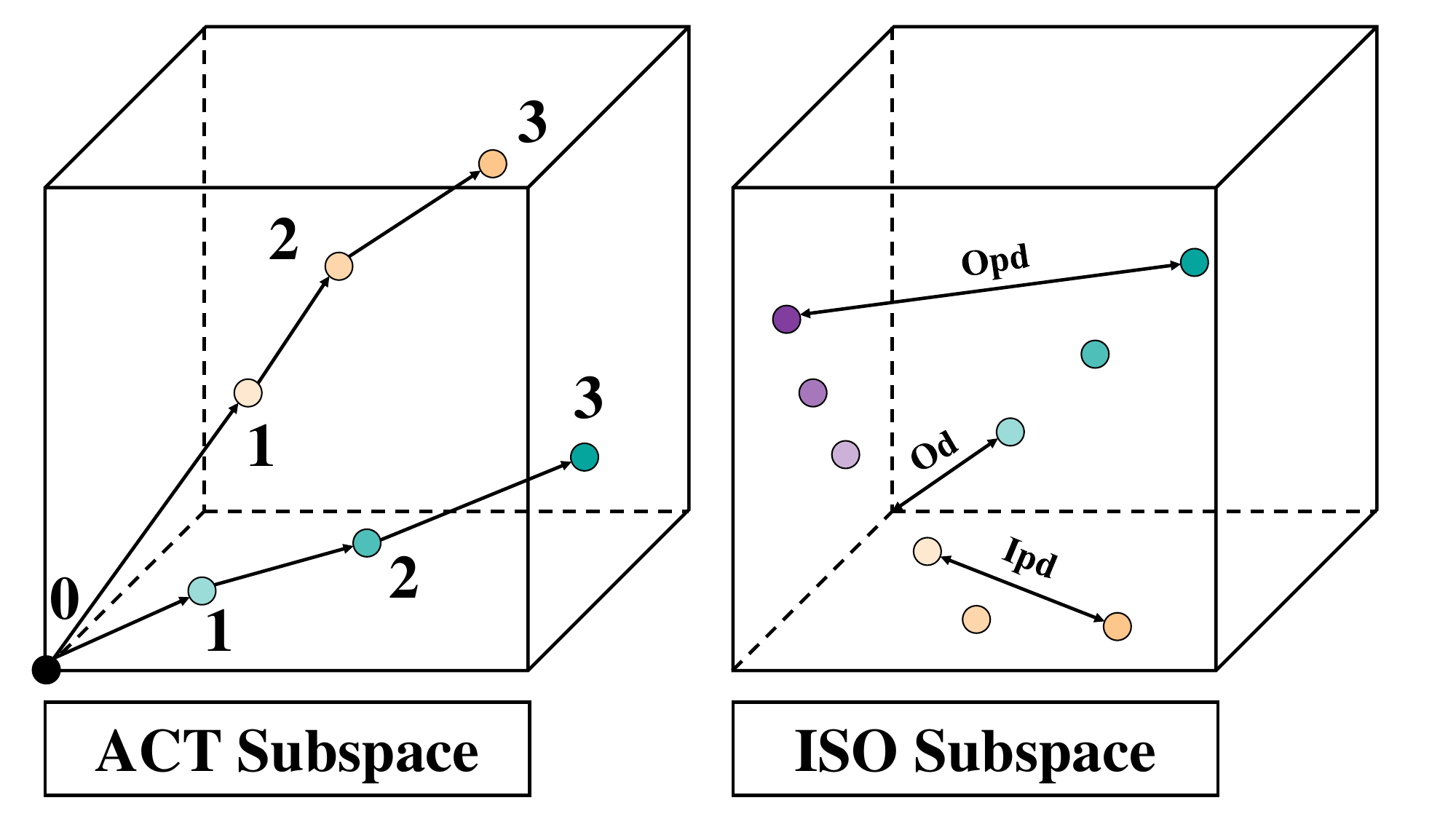} 
\caption{Subspaces of Linear Emotion Space}
\label{fig3}
\end{figure}\\\textbf{Action Subspace.} As shown in Fig.~\ref{fig3}, the space $\mathbb{E}$ comprises two subspaces: the Action Subspace $\mathbb{A}$ and the Isolation Subspace $\mathbb{I}$. 

Subspace $\mathbb{A}$ provides a representation for a unified AU level and emotion level by defining feature vectors for different emotional levels. The vector $\boldsymbol{u} \in \mathbb{A}$, as defined in Eqs.~\ref{Eq3} and \ref{Eq6}, primarily describes the actions of Facial Units. Considering the imbalanced distribution between AUs, we apply optimization function $\text{Opt}_{\text{1}}$:
\begin{equation}
\text{Opt}_{\text{1}}(AU_i) =\frac{AU_i - \mu_D}{\sigma_D} 
\label{Eq9}
\end{equation}
where $\mu_D$ and $\sigma_D$ are the mean and standard deviation of $AU_i$ based on the entire dataset. 
For each emotion, the Action Subspace defines three special $\boldsymbol{u}$ as feature vectors $\boldsymbol{uf}_{emo, level}$. The MEAD dataset classifies emotions excluding neutral into three base levels: $\text{level}_\text{1}\text{, level}_\text{2}\text{, and }\text{level}_\text{3}$, used as base level anchors. We individually collect $\boldsymbol{u}$ for each emotion at different levels to obtain $\boldsymbol{uf}_{emo, level}$:
\begin{equation}
\boldsymbol{uf}_{emo, k} = \frac{1}{N} \sum \boldsymbol{u}_{emo,level_k}
\label{Eq10}
\end{equation}
\begin{equation}
\boldsymbol{uf}_{neutral, 0} = \frac{1}{N} \sum \boldsymbol{u}_{neutral,level_1}
\label{Eq11}
\end{equation}

\begin{figure}[h]
\centering
\includegraphics[width=0.47\textwidth]{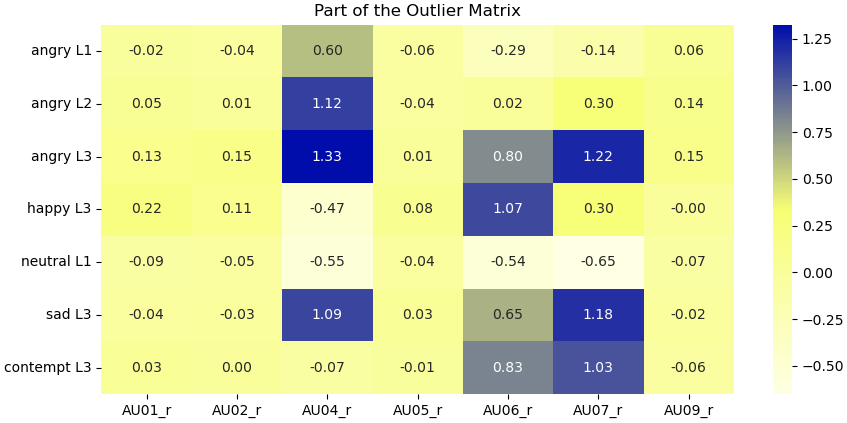} 
\caption{Part of the Outlier Matrix.}
\label{fig4}
\end{figure}

$\textbf{Effectiveness Proof}$ For a $\boldsymbol{uf_{emo, level}}$ to effectively capture the distinction from the general situation, it should exhibit outlier values. In the same video, the correlation between $\boldsymbol{u}$ decreases rapidly as the frame interval increases, whereas in different videos, $\boldsymbol{u}$ is independent. Belonging to the same AU, the $e_i$ is identically distributed. We apply the Central Limit Theorem and hypothesis testing to determine if an $e_i$ in $\boldsymbol{uf_{emo,int}}$ is an outlier. Given the parameters: $\mu = 0$, $\sigma = 1$, $n = 45,000$, confidence level = 0.999 (Z = 3.291). An $e_i$ is a significant anomaly if:
\begin{equation}
|e_i| > \frac{Z \cdot \sigma}{\sqrt{n}} = 0.0155
\label{Eq12}
\end{equation}
In the outlier matrix Fig.~\ref{fig4}, most AUs corresponding to $e_i$ are outliers. The effectiveness has been proven.\\\textbf{Isolation Subspace.} Subspace $\mathbb{I}$ provides a representation for subtle emotion transformations by isolating each emotional vector to capture information beyond AUs. 

The vector $\boldsymbol{v} \in \mathbb{I}$ is defined in Eq.~\ref{Eq7}. As defined in Eq.~\ref{Eq5}, elements $e_{35}$ to $e_{41}$ use a one-hot-like encoding to represent seven emotions excluding neutral, and $index$ corresponds to an emotion type. The origin distance is calculated as:
\begin{equation}
od = \sqrt{\sum_{i=18}^{34} {e_i^2}}
\label{Eq13}
\end{equation}
This distance $od$ reflects emotion level tendency, which we leverage to enable the network (CDAN) to learn additional representations. To unify the tendency expressed by AUs, we apply the optimization function $\text{Opt}_{\text{2}}$:
\begin{equation}
\text{Opt}_{\text{2}}(AU_i) = \frac{|AU_i|}{\sigma_{emo}}
\label{Eq14}
\end{equation}
where $\sigma_{emo}$ is the standard deviation of $AU_i$ for a specific emotion.

$\textbf{Isolation Proof}$
For a vector $\boldsymbol{v}$ to effectively represent the emotion level tendency, it must be distinguishable from vectors of other emotions. Poor results from Gaussian and k-means clustering (Silhouette Scores: -0.0217 and 0.0137) in $\mathbb{A}$ indicate the necessity of emotion isolation within $\mathbb{I}$.
Define both outer (Opd) and inner (Ipd) emotion distances:
\begin{equation}
d = \|\boldsymbol{v}_1 - \boldsymbol{v}_2\|
\label{Eq15}
\end{equation}
For $\boldsymbol{v}$ representing different emotions but with the same Od:
\begin{equation}
Ipd \leq od \times \sqrt{2} \leq Opd
\label{Eq16}
\end{equation}
This condition ensures that the network is always able to isolate vectors of different emotions as the level increases. The isolation has been proven.



\begin{figure}[h]
\centering
\includegraphics[width=0.45\textwidth]{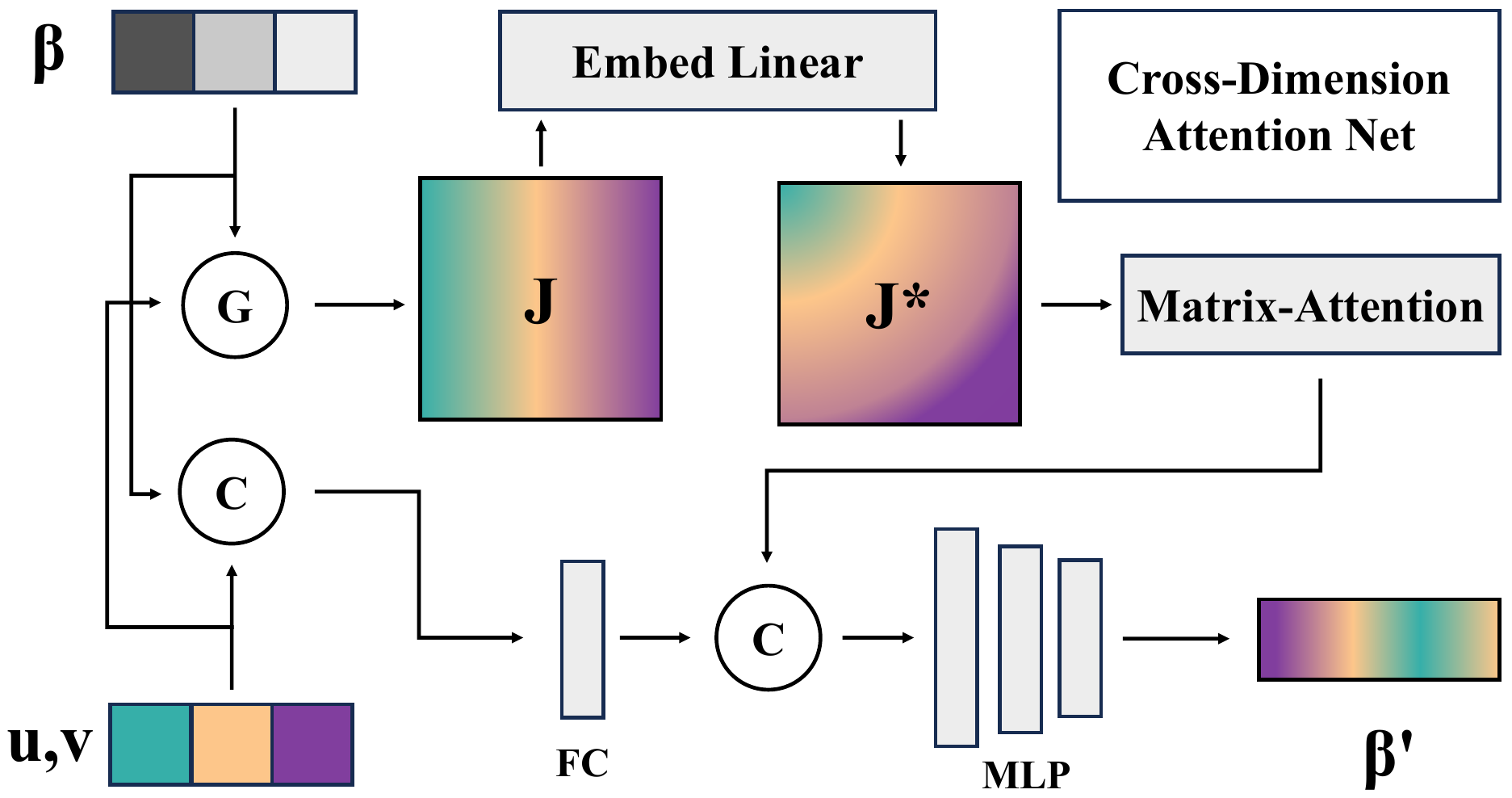} 
\caption{Structure of the Cross-Dimension Attention Net. Illustration in the process for a single frame's coefficients. \text{G} denotes the vector outer product operation, and \text{C} denotes vector concatenation. $\boldsymbol{u}$,$\boldsymbol{v}$ are vectors in ACT and ISO Subspace respectively. $\beta$ is 3DMM coefficients.}
\label{fig5}
\end{figure}
\subsection{Cross-Dimension Attention Net}
\label{sec:cdan}
Cross-Dimension Attention Net (CDAN) mines high-dimensional correlations between two low-dimensional inputs using matrix-attention and cross-dimension mechanism. The network inputs are two vectors, $\beta$ and either $\boldsymbol{u}$ or $\boldsymbol{v}$. Specifically, $\boldsymbol{u} \in \mathbb{A}$ is a $1 \times 17$ vector, and $\boldsymbol{v} \in \mathbb{I}$ is a $1 \times 24$ vector. The vector $\beta$, from Eq.~\ref{Eq1}, is a $1 \times 64$ vector.

From a cross-dimensional perspective, Action Units (AUs) affect 3DMM locations differently; for example, chin movements impact the lips more than the eyes. Different facial regions correlate to varying degrees. CDAN explores these correlations in higher and lower dimensions.

For higher dimensions, vectors $\boldsymbol{u}$ and $\boldsymbol{\beta}$ are combined into the Initial Joint Coefficient Matrix $\boldsymbol{J}$:

\begin{equation}
\boldsymbol{J} = \text{Combine}(\boldsymbol{u}, \boldsymbol{\beta})
\label{Eq17}
\end{equation}

Passing $\boldsymbol{J}$ through a linear layer produces the Combined Joint Coefficient Matrix $\boldsymbol{J^*}$:

\begin{equation}
\boldsymbol{J^*} = \text{Linear}(\boldsymbol{J})
\label{Eq18}
\end{equation}

Here, $\boldsymbol{J^*}$ is shaped $17 \times 64$. The 17 dimensions represent facial units as sequence length, and the 64 dimensions represent 3DMM coefficients as embedding vectors.  A matrix-attention mechanism adds to correlation mining.

For lower dimensions, concatenate $\boldsymbol{u}$ and $\beta$:

\begin{equation}
\text{Concat}(\boldsymbol{u}, \boldsymbol{\beta}) \rightarrow \text{FC}(\cdot)
\end{equation}

Eq.~\ref{Eq20} describes parallel inputs to predict transformed 3DMM coefficients:

\begin{equation}
\beta' = \text{MLP}(\text{FC}(\boldsymbol{u}, \beta), \text{Att}(\boldsymbol{u}, \beta))
\label{Eq20}
\end{equation}

As shown in Fig.~\ref{fig2}, utilizing $\boldsymbol{{w}^{'}} \in \mathbb{E}$ from the Emo Injector, our method employs two CDANs in series and parallel to achieve controllable deformation of a facial 3D model. Each CDAN receives vectors from the ACT and ISO subspaces, respectively. For finer representation in the ISO subspace, when $\boldsymbol{v}$ is an input, the other input is the predicted coefficients $\beta'$ obtained from $\boldsymbol{u}$ and $\beta$.
\begin{figure}[h]
\centering
\includegraphics[width=0.42\textwidth]{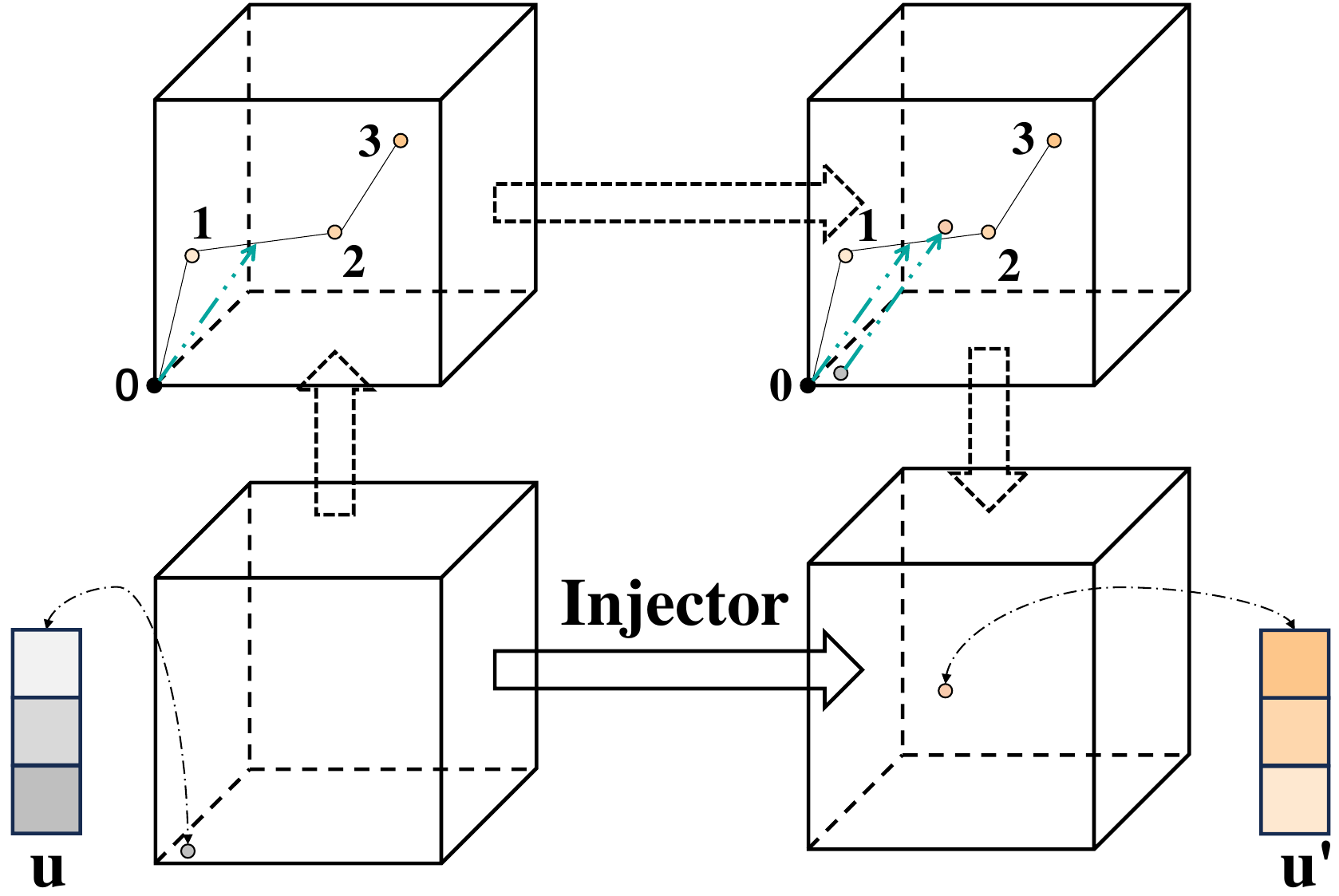} 
\caption{The Emotion Injector.}
\label{fig6}
\end{figure}

\begin{figure*}[h]
\centering
\includegraphics[width=\textwidth]{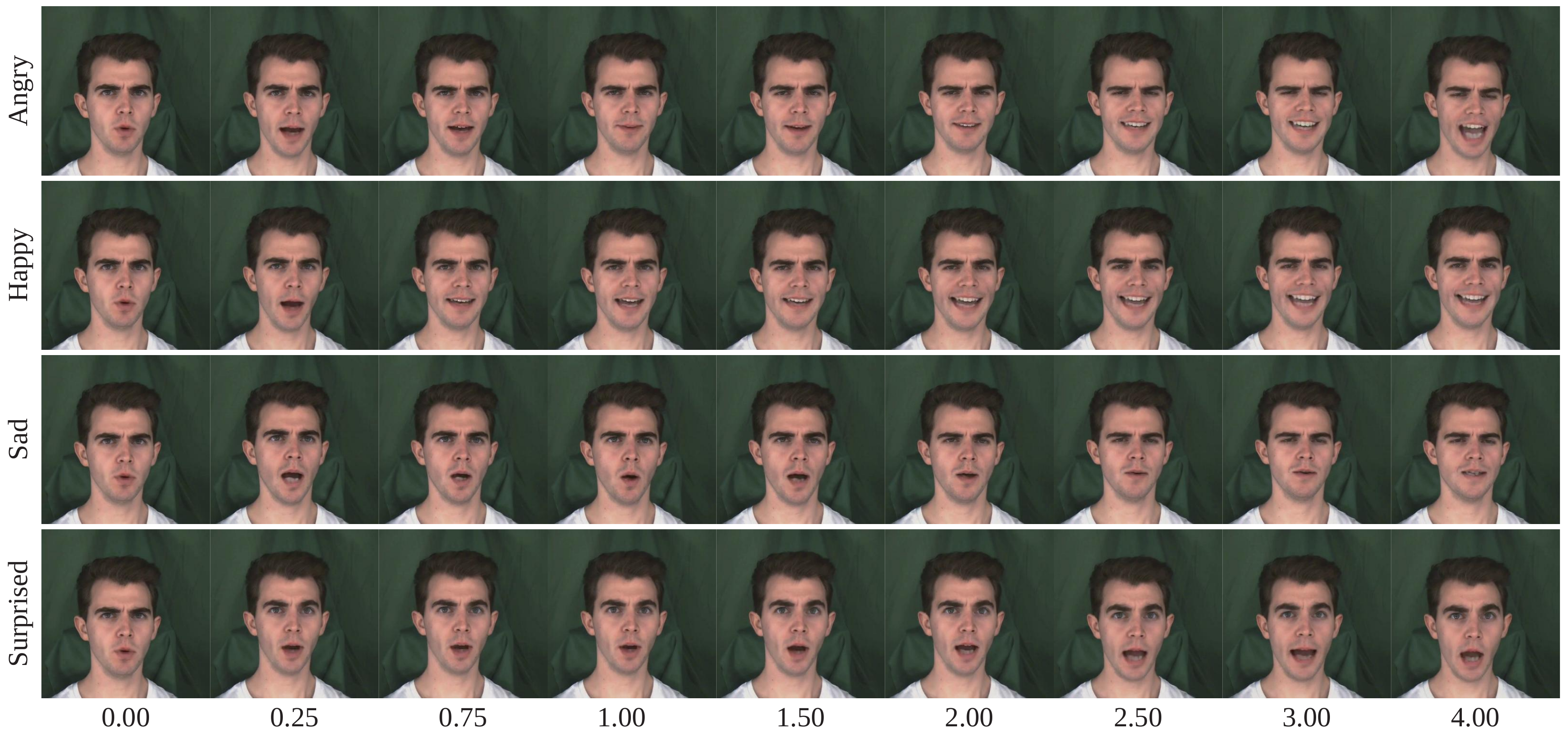} 
\caption{Fine-grained emotion editing.}
\label{fig7}
\end{figure*}


\subsection{Other Key Components}
\label{sec:other key components}
\textbf{Linear Emotion Space Recon.} The reconstruction of emotion utilizes OpenFace \cite{baltruvsaitis2016openface} to extract the origin AUs sequence from a given AU source and then processes the AUs sequence, mapping it to the ACT and ISO subspaces. This involves deriving $\mu_D$, $\sigma_D$, and each $\sigma_{emo}$ statistically before training, as shown in Eq.~\ref{Eq9} and Eq.~\ref{Eq14}. \\\textbf{Audio Encoder-Decoder.} Considering that it is not always easy for users to provide AU source video in actual use, our method can be video-driven as well as audio-driven. Shown in Fig.~\ref{fig2}, an Audio Encoder, based on a ResNet model \cite{zhang2023sadtalker}, encodes audio inputs into content vectors for further collaboration with Decoders. AU Decoder provides generative input for vector $\boldsymbol{w} \in \mathbb{E}$ when AU inputs are absent. Offset Decoder enhances the predicted 3D coefficients by incorporating content vectors and leveraging predictions from another pretrained audio encoder \cite{zhang2023sadtalker}. Since audio features contain unique information, Offset Decoder compensates for residuals in the 3D coefficients after CDAN prediction.
\\\textbf{Emotion Injector.} The Emotion Injector, based on Fig.~\ref{fig4}, Eq.~\ref{Eq9}, and Eq.~\ref{Eq10}, utilizes twenty-two feature vectors $\boldsymbol{uf}$ for eight emotions, stored statistically before testing.  $\boldsymbol{uf}_{neutral,0}$ is defined where emotion level is 0. $\boldsymbol{u}$ is updated first and then  $\boldsymbol{v}$ is updated. As shown in Fig.~\ref{fig6} and based on Eq.~\ref{Eq8}, when user's control label is $(emo, level)$, the transformation is as follows:
\begin{equation}
\begin{array}{c}
i = \lceil level \rceil \quad\quad\quad  j = \lfloor level \rfloor
\end{array}
\end{equation}
\begin{equation}
\begin{aligned}
\boldsymbol{uf}_{emo, level} = \, & (\boldsymbol{uf}_{emo, i} - \boldsymbol{uf}_{emo, j}) \\
&\cdot (level - j) + \boldsymbol{uf}_{emo, j}
\end{aligned}
\label{Eq19}
\end{equation}
\begin{equation}
\begin{array}{c}
\boldsymbol{u}_{inj} = \boldsymbol{uf}_{emo,level} - \boldsymbol{uf}_{neutral,0}
\end{array}
\end{equation}
\begin{equation}
\begin{array}{c}
\boldsymbol{u'} = \boldsymbol{u}_{inj} + \boldsymbol{u}
\end{array}
\label{Eq21}
\end{equation} \\\textbf{3D-Aware Face Render.} Inspired by Sadtalker \cite{zhang2023sadtalker}, we use identity, pose, and transformed 3D coefficients to generate a video with a pre-trained renderer.

\section{Experiment}
In this section, we demonstrate three key points through multiple experiment settings: 
(1) the adaptation to Linear Emotion Space (LES) definition enables LES-Talker to achieve fine-grained emotion editing; (2) the improved visual achieved by LES-Talker; (3) the effective design of the CDAN and other key components.

\begin{figure*}[h]
\centering  
\includegraphics[width=\textwidth]{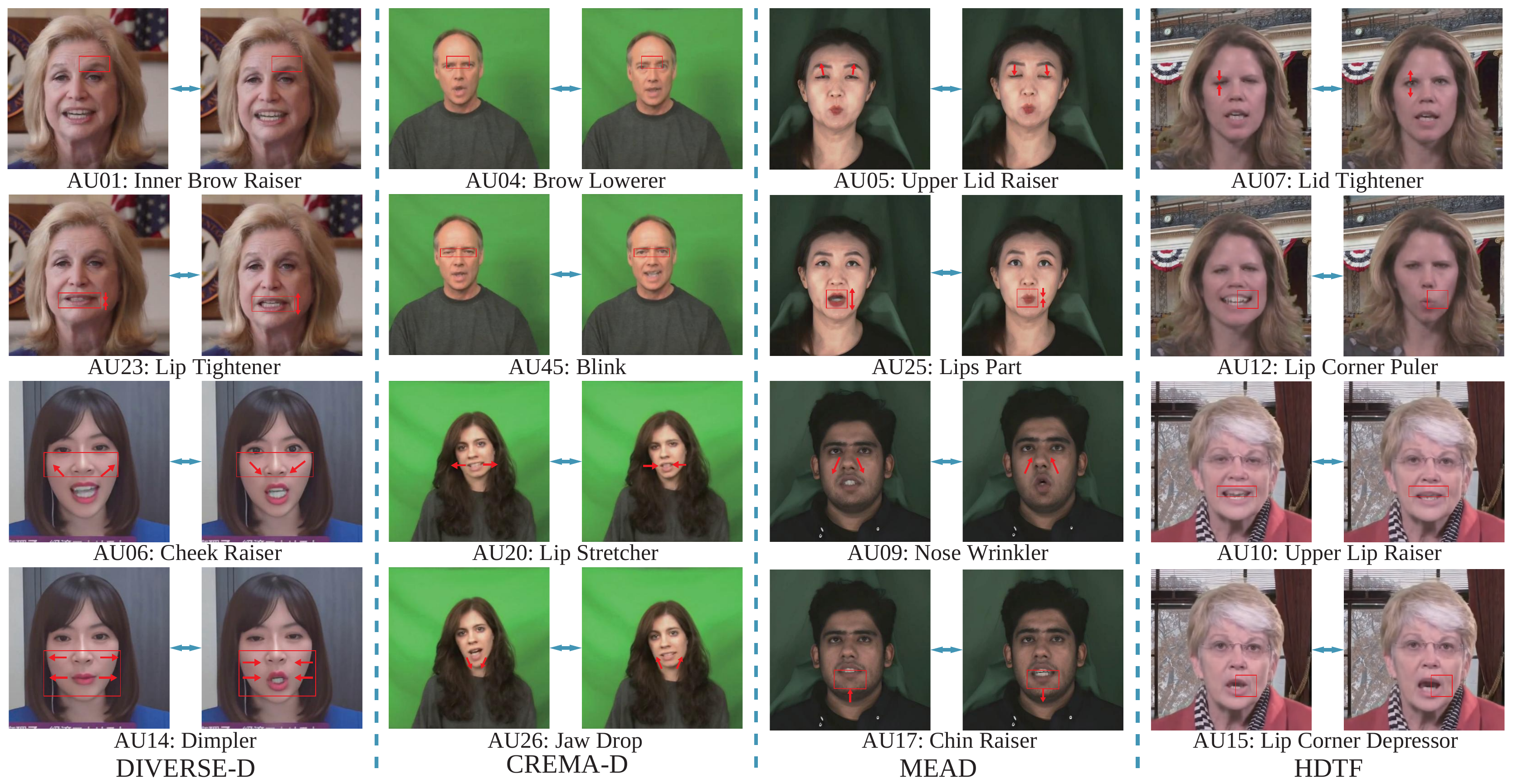} 
\caption{Fine-grained AUs editing.}
\label{fig8}
\end{figure*}

\subsection{Experimental Settings.}
\textbf{Datasets.} We utilized the MEAD \cite{wang2020mead} as our primary dataset. To demonstrate the generalizability of our method, we also conducted evaluations using the CREMA-D \cite{cao2014crema}, HDTF \cite{zhang2021flow}, and a diverse dataset \cite{zhang2024comparative}.\\\textbf{Implementation Details.} We performed all experiments on an NVIDIA RTX 3090Ti GPU using the PyTorch platform. The input audio was sampled at 16,000 Hz and represented as a 0.2-second mel spectrogram. Videos were cropped and resized to 512\(\times\)512. OpenFace \cite{baltruvsaitis2016openface} and DeepFace3DReconstruction \cite{deng2019accurate} were used to extract AU values and 3DMM coefficients from input images. Our training employs a two-step coarse-to-fine strategy. We provide more details about training in the supplementary materials.
\\\textbf{Evaluation Metrics.} Video quality was assessed using the following four metrics: Frechet Inception Distance (FID), Structural Similarity (SSIM), Peak Signal-to-Noise Ratio (PSNR), Cumulative Probability Blur Detection (CPBD). Lip synchronization was evaluated with three metrics using Syncnet \cite{chung2017out}: Lip Sync Confidence (AVConf), Lip Offset (AVOffset), and Minimum Offset (MinDist).

\begin{figure}[t]
\centering
\includegraphics[width=0.45\textwidth]{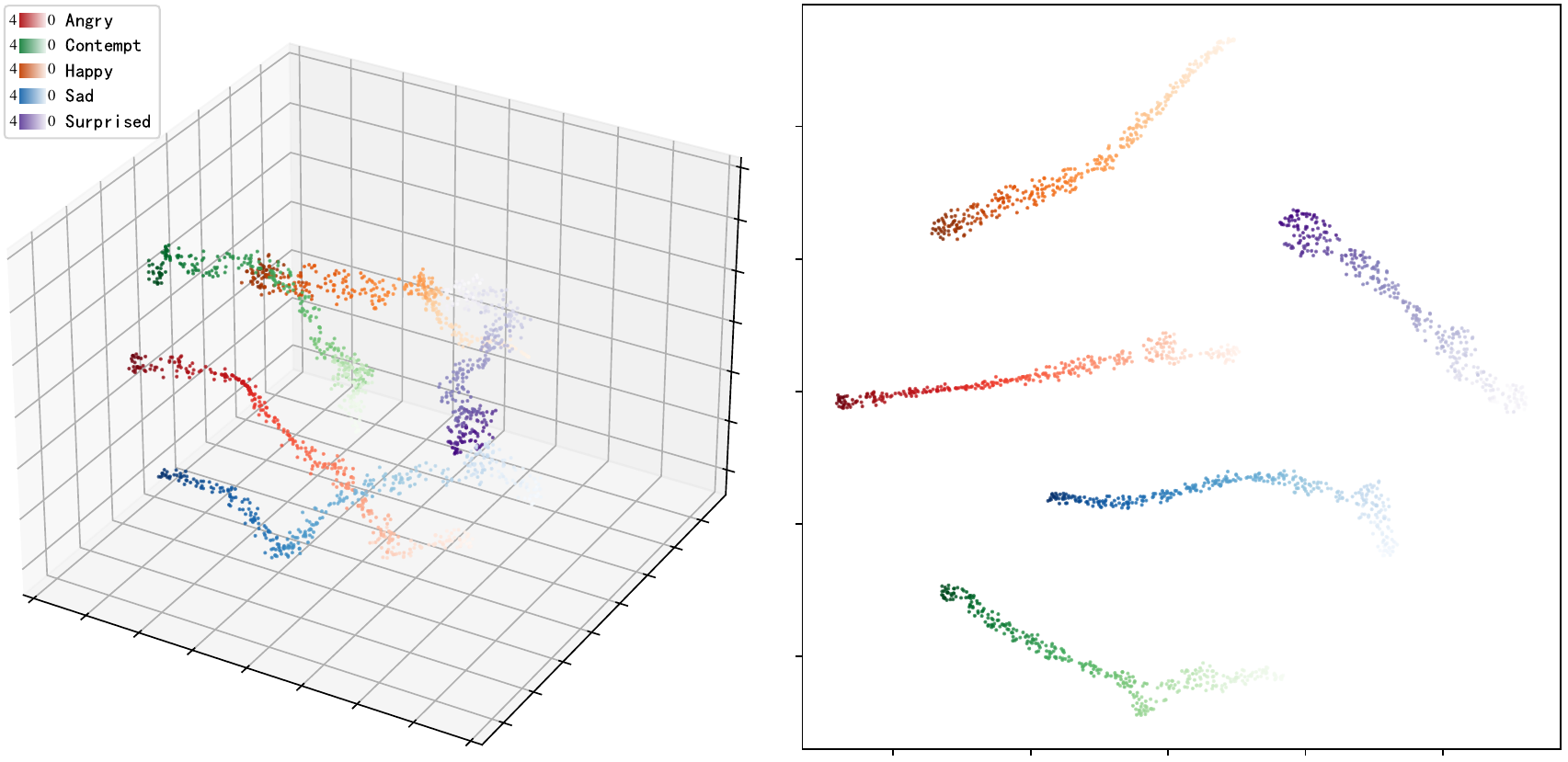} 
\caption{t-SNE Visualization. }
\label{tsne}
\end{figure}

\subsection{Editing Effectiveness}
We demonstrate the effectiveness of the editing by achieving the expected experimental results.
\\\textbf{Fine-grained emotion editing.} In the test set, we selected identity images and AU sources from neutral emotion videos, along with the corresponding audio. The Emo Injector method was strictly followed to inject different emotions at various levels. We compared 
frames with the same emotion but varying levels. In Fig.~\ref{fig7}, the horizontal axis represents the level, while the vertical axis represents different emotions. We achieved fine-grained editing for levels 0-3 and effective editing for hyperdomain levels (level $>$ 3).
\\\textbf{Fine-grained AUs editing} As shown in Fig.~\ref{fig1}, we can set the level in a arbitrary values. To validate the model's generalization and effectiveness, we imposed strict conditions: absent AU source, cross-dataset, and cross-identity. Control conditions involved injecting +2.5 and -2.5 into the AUs in the ACT Subspace, with each image pair having +2.5 on the left and -2.5 on the right. As illustrated in Fig.~\ref{fig8}, our method achieves editing control over nearly all defined AUs. \\\textbf{Space Visualization.} We generated 1,755 videos across five emotions, with levels from 0 to 4 in steps of 0.0114. Each video's frames were transformed into 41-dimensional LES representations and averaged, followed by t-SNE for dimensionality reduction. In Fig.~\ref{tsne}, we can observe 
that the spatial representation reveals a clear emotional gradient and distinct separation between emotions.
\begin{table*}[]
\setlength{\tabcolsep}{1.5mm}
\footnotesize
\centering
\begin{tabular}{l c c c c c c c c c c c}
\toprule
\multirow{2}{*}[-0.25em]{\centering Method} & \multicolumn{4}{c}{Video Quality Comparison} & \multicolumn{3}{c}{Lip Synchronization Comparison} & \multicolumn{4}{c}{User Study} \\ \cmidrule(lr){2-5} \cmidrule(lr){6-8} \cmidrule(lr){9-12}
 & FID↓ & SSIM↑ & PSNR↑ & CPBD↑ & MinDist↓ & AVConf↑ & AVOffset(→0) & LipSyn↑ & EmoAcc↑ & Reality↑ & Quality↑ \\ \midrule
Real Video & 0.000 & 1.000 & 31.734 & 0.265 & 7.869 & 6.564 & -2.000 & 4.54 & 4.74 & 4.88 & 4.89 \\  
MEAD\cite{wang2020mead} & 146.454 & 0.469 & 14.864 & 0.191 & 11.957 & 2.674 & -2.000 & 2.73 & 3.42 & 3.87 & 3.72 \\ 
EVP\cite{ji2021audio} & 56.650 & 0.453 & 16.308 & \textbf{0.341} & 12.443 & 3.163 & 5.000 & 2.83 & 3.55 & 3.71 & 3.85 \\ 
EAMM\cite{ji2022eamm} & 204.002 & 0.396 & 12.832 & 0.135 & 10.091 & 3.046 & -4.000 & 3.12 & 3.72 & 3.65 & 3.56 \\ 
SadTalker\cite{zhang2023sadtalker} & 25.746 & 0.718 & 20.728 & 0.249 & \textbf{8.875} & 6.484 & 1.000 & \textbf{3.71} & - & 3.30 & 3.86 \\
EAT\cite{gan2023efficient} & 135.439 & 0.706 & 13.139 & 0.228 & 7.531 & \textbf{8.164} & -2.000 & 3.38 & 3.76 & 3.91 & 3.93 \\ \midrule
Ours & \textbf{24.783} & \textbf{0.743} & \textbf{21.686} & 0.261 & 9.800 & 7.284 & \textbf{0.000} & 3.59 & \textbf{3.79} & \textbf{3.96} & \textbf{4.10} \\
\bottomrule
\end{tabular}
\caption{Quantitative comparisons on video quality, lip synchronization, and user study.}
\label{comparision_combined}
\end{table*}

\begin{figure}[h]
\centering
\includegraphics[width=0.40\textwidth]{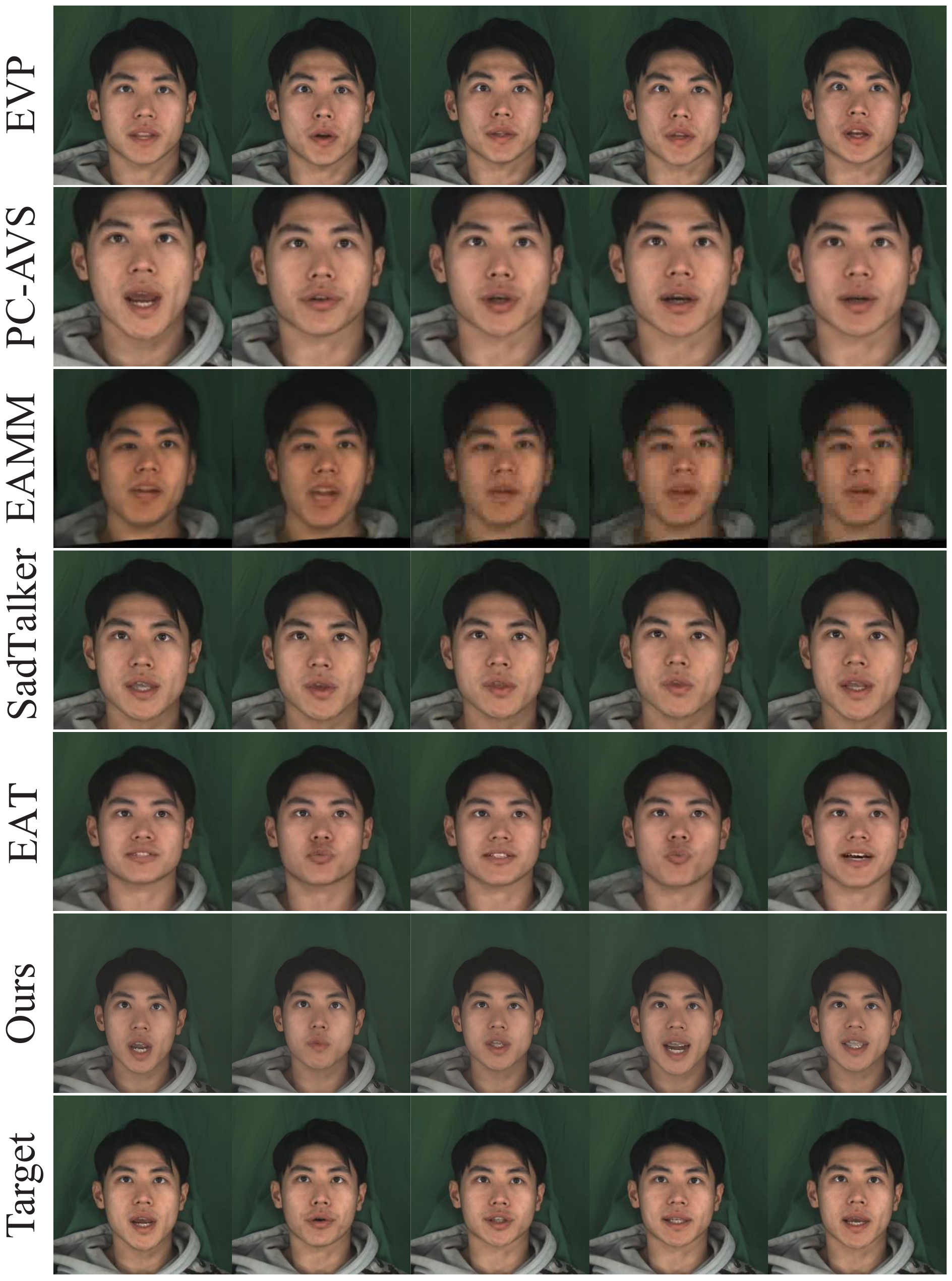} 
\caption{Visual quality comparison.}
\label{fig9}
\end{figure}

\subsection{Visual Quaility}

We compare our method with state-of-the-art open-source emotion-driven models (EAMM, EVP, EAT) and the one-shot method SadTalker, achieving superior visual quality.
\begin{table}[h]
\setlength{\tabcolsep}{3.8mm}
\footnotesize 
\centering
\begin{tabular}{l c c c c c} 
\toprule
\multirow{2}{*}[-0.25em]{\centering Emo Type} & \multicolumn{5}{c}{ Emo Level} \\ 
\cmidrule(lr){2-6}
 & 0.50 & 1.00 & 1.50 & 2.00 & 2.50  \\ 
\midrule
Angry & 0.87 & 1.20 & 1.60 & 1.80 & 1.84 \\ 
Happy & 0.83 & 1.41 & 1.77 & 2.00 & 2.23 \\ 
Sad & 1.22 & 1.42 & 1.51 & 2.07 & 2.07 \\ 
Surprised & 1.73 & 1.92 & 2.01 & 2.04 & 2.28 \\ 
\midrule
Average & 1.16 & 1.49 & 1.72 & 1.98 & 2.10 \\ 
\bottomrule
\end{tabular}
\caption{User-perceived level.}
\label{tab2}
\end{table}\\

\begin{table}[b]
\setlength{\tabcolsep}{1.95mm}
\footnotesize
\small
\centering
    \begin{tabular}{c c c c c c c}
        \toprule
        Opt. & OD & AS & MinDist$\downarrow$ & AVConf$\uparrow$ & AVOffset($\to 0$) \\
        \midrule
         &  &  & 13.184 & 2.081 & -11.000 \\
        & \checkmark &  & 12.250 & 2.911 & -1.000 \\
        &  & \checkmark & 10.727 & 6.330 & 0.000 \\
        \checkmark &  & \checkmark & 10.403 & 6.563 & 0.000 \\
          & \checkmark & \checkmark  &  10.152 & 6.876 & 0.000 \\
        \checkmark & \checkmark & \checkmark &  \textbf{9.800} & \textbf{7.284} & \textbf{0.000} \\
        \bottomrule
    \end{tabular}

\caption{Quantitative comparisons in ablation study. }
\label{tab4}
\end{table}

\begin{figure}[t]
\centering
\includegraphics[width=0.47\textwidth]{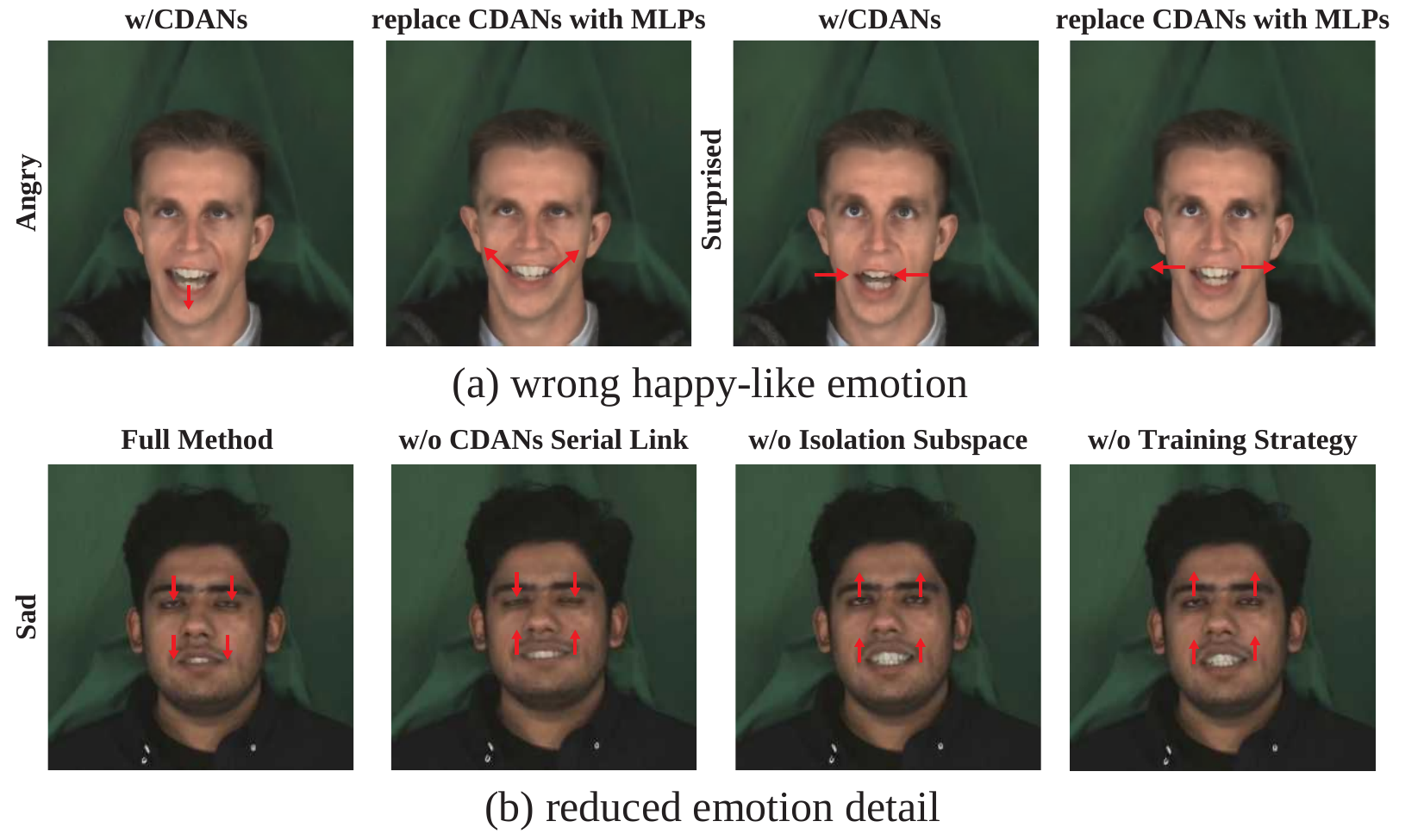} 
\caption{Generation in ablation study. }
\label{fig10}
\end{figure}

\textbf{User Study.} We conducted user studies with 15 participants to evaluate our model across multiple metrics (rated 1–5) compared to other methods, as detailed in Tab.\ref{comparision_combined}. Our method outperformed others. As illustrated in Tab.\ref{tab2}, we assessed user-perceived emotion levels when presented with videos labeled $(emo, level)$. The results confirm a consistent upward trend between theoretical and perceived emotion intensity, validating our framework’s effectiveness. Meanwhile, intensity shifts in sadness and surprise are subtler than in happiness and anger.

\subsection{Ablation Study.}
We conducted ablation studies in multiple conditions to comprehensively demonstrate the contribution. Additionally, the absence of serial connection between two-level CDANs leads to poor performance. The incomplete methods shown in Fig.~\ref{fig10} lose the ability to express distinct emotional details, causing angry, surprised, and sad to resemble happy-like emotions, primarily by widening the mouth as the emotion level increases.  Opt., OD and AS represents the mechanisms referred to Eq.~\ref{Eq9} and Eq.~\ref{Eq14} in LES Recon., Offset Decoder and Au Source respectively. We display progressively incorporating components in Tab.~\ref{tab4}. More details are in the supplementary materials.

\section{Conclusion}
In this paper, we propose Linear Emotion Space (LES), a fine-grained emotion definition capturing facial unit actions and subtle details beyond AUs, along with a fine-grained emotion editing model, LES-Talker. We use 3DMM as an intermediate representation and propose Cross-Dimension Attention Net to guide the controllable deformation of the 3D model. We also design various key components to adapt multi-modal features. LES-Talker delivers high visual quality and fine-grained emotion editing, performing diverse tasks. Our theoretical definition of LES distinguishes the model and could serve as a foundation for future research.
{
    \small
    \bibliographystyle{ieeenat_fullname}
    \bibliography{main}
}
\clearpage

\appendix
\section*{Appendix} 
\section{Further Description of LES}
\begin{table}[h]
\small
\centering
\begin{tabular}{|c|c||c|l|}
\hline
\textbf{AUs} & \textbf{Action Description} & \textbf{AUs} & \textbf{Action Description} \\ \hline
AU1                              & Inner Brow Raiser&AU14          & Dimpler                   \\ \hline
AU2                              & Outer Brow Raiser&AU15          & Lip Corner Depressor      \\ \hline
AU4                              & Brow Lowerer     &AU17          & Chin Raiser               \\ \hline
AU5                              & Upper Lid Raiser &AU20          & Lip Stretcher             \\ \hline                                
AU6                              & Cheek Raiser     &AU23          & Lip Tightener             \\ \hline
AU7                              & Lid Tightener    &AU25          & Lips Part                  \\ \hline
AU9                              & Nose Wrinkler    &AU26          & Jaw Drop                   \\ \hline
AU10                             & Upper Lip Raiser &AU45          & Blink                      \\ \hline 
AU12                             & Lip Corner Puller&              &                             \\ \hline
\end{tabular}
\caption{Facial Action Units Description.}
\label{Tab1}
\centering
\end{table}
\subsection{Physical Meaning}
Our proposed \textit{Linear Emotion Space} contains a total of 41 dimensions. The first 17 dimensions, which make up the \textit{Action Subspace} \(\mathbb{A}\), directly correspond to the 17 Facial Action Units (AUs) listed above. Their values have a unified physical meaning across all emotions, representing the amplitude of movement of each facial unit. The next 17 dimensions (18-34) form the \textit{Isolation Subspace} \(\mathbb{I}\) and also correspond to the same 17 AUs. However, in this subspace, the values have distinct physical meanings under each emotion, representing the magnitude of fluctuation in the movement of each facial unit. Finally, the last 7 dimensions (35-41) of the \textit{Isolation Subspace} are dedicated to isolating individual emotions and directly capturing the overall intensity of each emotion.
\begin{table}[h]
\centering
\begin{tabular}{|c|l|c|}
\hline
\textbf{Cluster} & \textbf{Top Emotion-Level Pairs} & \textbf{Percentage} \\ \hline
0 & happy\_level\_3 (46.67\%)& 4.55 \\ \hline
1 & contempt\_level\_3 (32.40\%)& 10.81 \\ \hline
2 & disgusted\_level\_3 (42.24\%)& 7.81 \\ \hline
3 & fear\_level\_3 (39.32\%) & 7.88 \\ \hline
4 & angry\_level\_3 (25.85\%)& 22.66 \\ \hline
5 & neutral\_level\_1 (48.41\%)& 27.54 \\ \hline
6 & sad\_level\_3 (37.66\%)& 10.37 \\ \hline
7 & happy\_level\_3 (68.67\%)& 8.38 \\ \hline
\end{tabular}
\caption{GMM Clusters for vectors in ACT Subspace.}
\label{tab:GMM}
\end{table}

\begin{table}[h]
\centering
\begin{tabular}{|c|l|c|}
\hline
\textbf{Cluster} & \textbf{Top Emotion-Level Pairs} & \textbf{Percentage} \\ \hline
0 & neutral-level\_1 (42.63\%)& 33.80 \\ \hline
1 & sad-level\_3 (38.66\%)& 13.06 \\ \hline
2 & happy-level\_3 (59.41\%)& 3.40 \\ \hline
3 & sad-level\_3 (23.49\%)& 5.59 \\ \hline
4 & contempt-level\_3 (27.73\%)& 12.63 \\ \hline
5 & contempt-level\_3 (40.43\%)& 12.66 \\ \hline
6 & disgusted-level\_3 (42.16\%)& 9.66 \\ \hline
7 & happy-level\_3 (64.10\%)& 9.19 \\ \hline
\end{tabular}
\caption{KMeans Clusters for vectors in ACT Subspace.}
\label{tab:kmeans}
\end{table}
\subsection{Cluster Analysis}
As shown in Tab.~\ref{tab:GMM} and Tab.~\ref{tab:kmeans}, we provide a detailed demonstration of the clustering analysis mentioned in the main text. When performing emotion reconstruction using only the \textit{Action Subspace}, both K-means and GMM clustering exhibit poor results. Within each cluster, the most frequent emotion-level pair does not significantly dominate. Across clusters, the sample points of each class—which should be approximately equal in number—are unevenly distributed, and the different emotions cannot be effectively separated.
\begin{figure*}[h]
\centering
\includegraphics[width=0.97\textwidth]{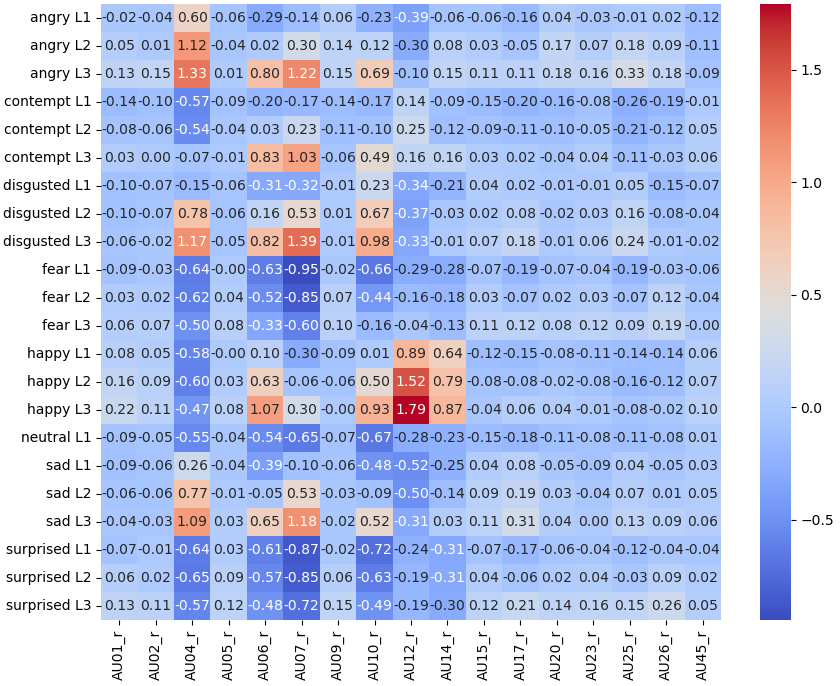} 
\caption{The complete outlier matrix.}
\label{fig:fulloutliermarix}
\end{figure*}
\subsection{Complete Outlier Matrix}
As illustrated in Fig.~\ref{fig:fulloutliermarix}, we present the complete outlier matrix. The horizontal axis corresponds to the indices of the 17 controllable action facial units, while the vertical axis denotes the emotion-level pairs. The outlier matrix is precomputed prior to training using the LES definitions described in the main text, leveraging a pre-trained model and statistical methods. The matrix provides 22 feature vectors $\mathbf{uf}$ in the Action Subspace, which are employed for emotion transformation in the Emotion Injector.

\begin{figure*}[h]
\centering
\includegraphics[width=0.97\textwidth]{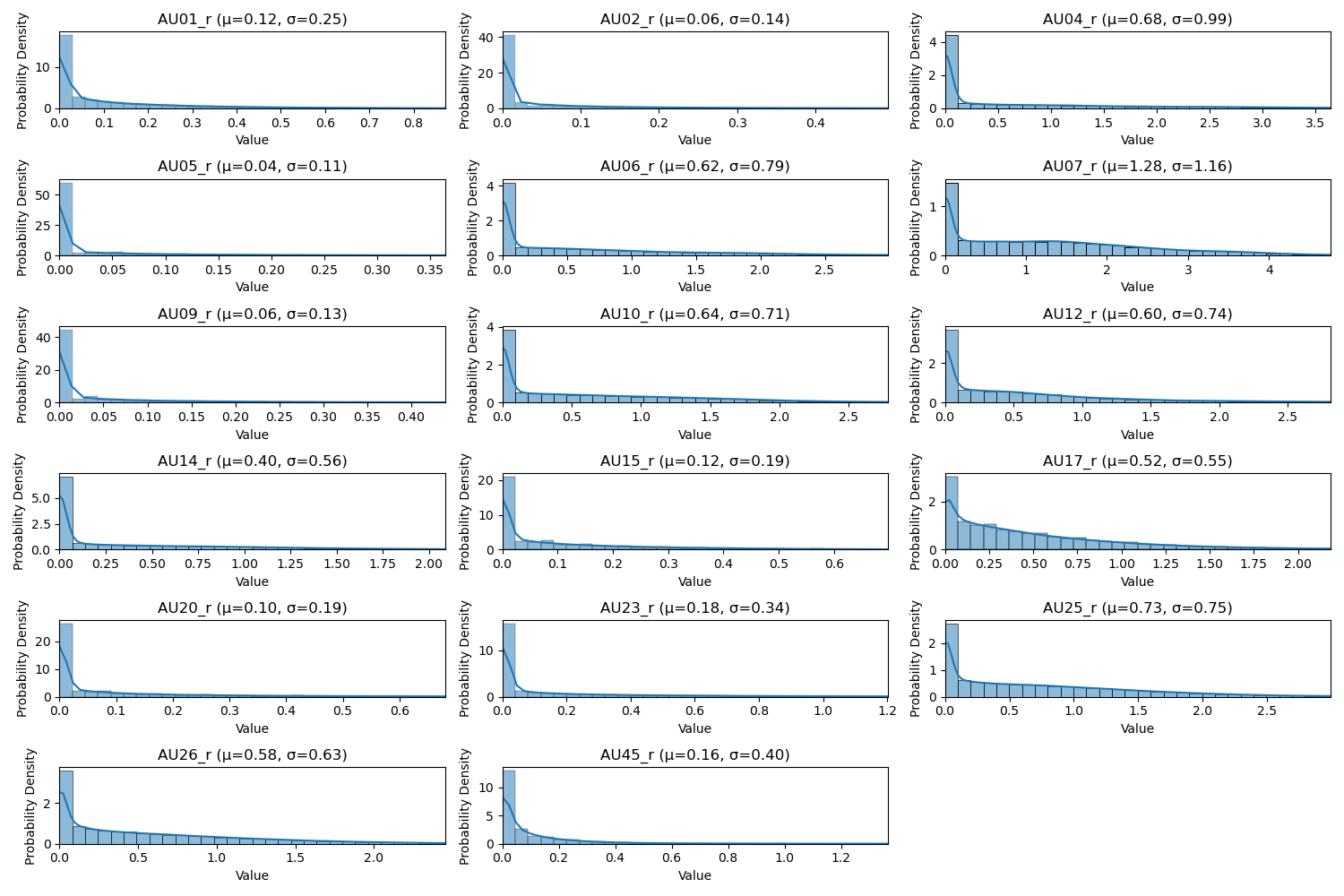} 
\caption{Unbalanced distribution before standardization.}
\label{fig:unbalanced_distribution}
\end{figure*}
\begin{figure*}[htbp]
\centering
\includegraphics[width=0.97\textwidth]{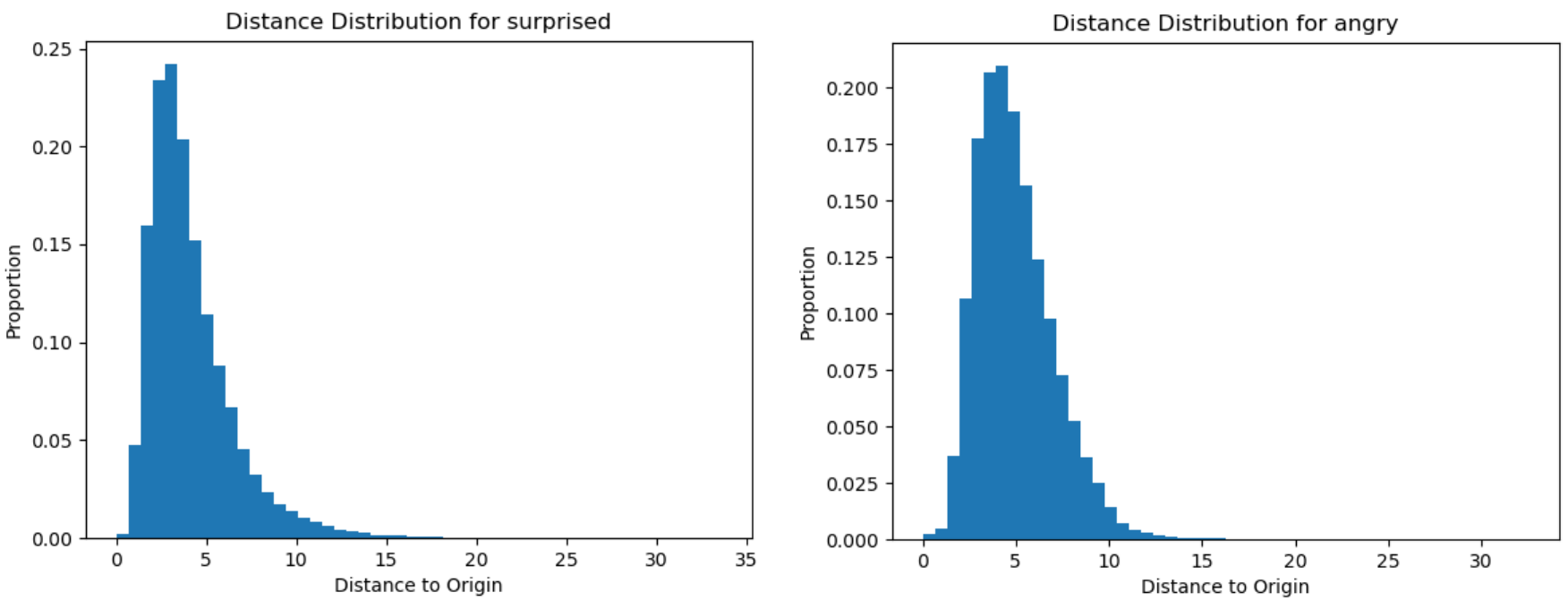} 
\caption{Od distribution in ISO Subspace.}
\label{fig:od}
\end{figure*}
\subsection{Distribution Optimizing}
As shown in Fig.~\ref{fig:unbalanced_distribution}, prior to applying the Opt1 and Opt2 methods defined in the LES, all AUs exhibit an unbalanced distribution. This is partly due to inherent limitations of the dataset itself, and partly due to defects in the AU extraction tools. Such an unbalanced distribution leads to the inability to establish a unified representation in the subsequent LES-defined space, and also results in deficiencies in the final visual effects. As illustrated in Fig.~\ref{fig:od}, after distribution optimization, in the \textit{Isolation Space}, the vector distance $Od$ used to represent emotional levels achieves a uniform distribution across different emotions, providing support for a unified level representation of different emotions.

\section{Training Strategy}
\subsection{Coarse-to-Fine Strategy}
During training, a two-step strategy from coarse to fine was employed as follows:

\textbf{Coarse Training}: The coarse training aligns with the design concept of the \textit{Action Subspace}, maximizing the use of AUs for facial action representation. Train the first-level CDAN individually using the first frame of each video as the reference image with \(\boldsymbol{u} \in \mathbb{A}\). Since the first frame better matches the identity information of subsequent frames, this training approach provides more precise facial action control.

\textbf{Fine Training}: As using only AUs is insufficient, facial expressions beyond AUs need to be further adjusted through the \textit{Isolation Subspace} in the fine training. Freeze the first CDAN, then train the second-level CDAN using the neutral emotion frame as the reference image with \(\boldsymbol{v} \in \mathbb{I}\). Because the final goal is to perform arbitrary emotion transformations starting from a neutral emotion, our method enables the network to have stronger emotion transformation capabilities.

Finally, freeze both levels of CDAN and train the Offset Decoder separately, utilizing information from the audio modality to compensate for the residuals between the remaining predicted coefficients and the ground truth coefficients. This specialized training method not only allows the model to achieve the desired emotional expression capabilities but also attains higher visual quality.

\subsection{Hyper-parameters}
To ensure rapid and stable convergence of the model training towards lower loss values, we employed the Adam optimizer and conducted extensive hyperparameter tuning, including adjustments to the factor and patience parameters. Experimental results demonstrated that the initial learning rate for each epoch was set as \( lr \times decay^{\text{epoch}} \), where \( lr \in \{10^{-3}, \mathbf{10}^{-4}, 10^{-5}\} \) and \( decay \in \{0.9, \mathbf{0.86}, 0.74\} \). The batch size was configured as \( \{\mathbf{10}, 20, 30\} \), and patience was set to \( \{1500, \mathbf{15000}, 30000\} \). Larger batch sizes and excessively low learning rates resulted in prolonged training times, while larger factors and learning rates compromised training stability. Additionally, smaller decay rates and patience values led to insufficient convergence. Training was terminated after 30 epochs.


\setlength{\tabcolsep}{6mm}
\begin{table}[h]
\small
\centering
{
\begin{tabular}{l c c}
\toprule
Guide Space& Method & FLD↓ \\
\hline
ACT & MLP  & 1.959  \\
ACT & CDAN & \textbf{1.150} \\
ISO & MLP  & 2.230 \\
ISO & CDAN w/o serial & 2.012\\
ISO & CDAN & \textbf{1.421} \\ 
\bottomrule
\end{tabular}}
\caption{Facial Landmark Distance (FLD) under emotional conditions, based on 3DMM, summed over 200 frames.}
\label{cdan-ab}
\end{table}
\section{Further Experiment}
\subsection{Facial Landmark Distance}
\begin{figure*}[h]
\centering
\includegraphics[width=\textwidth]{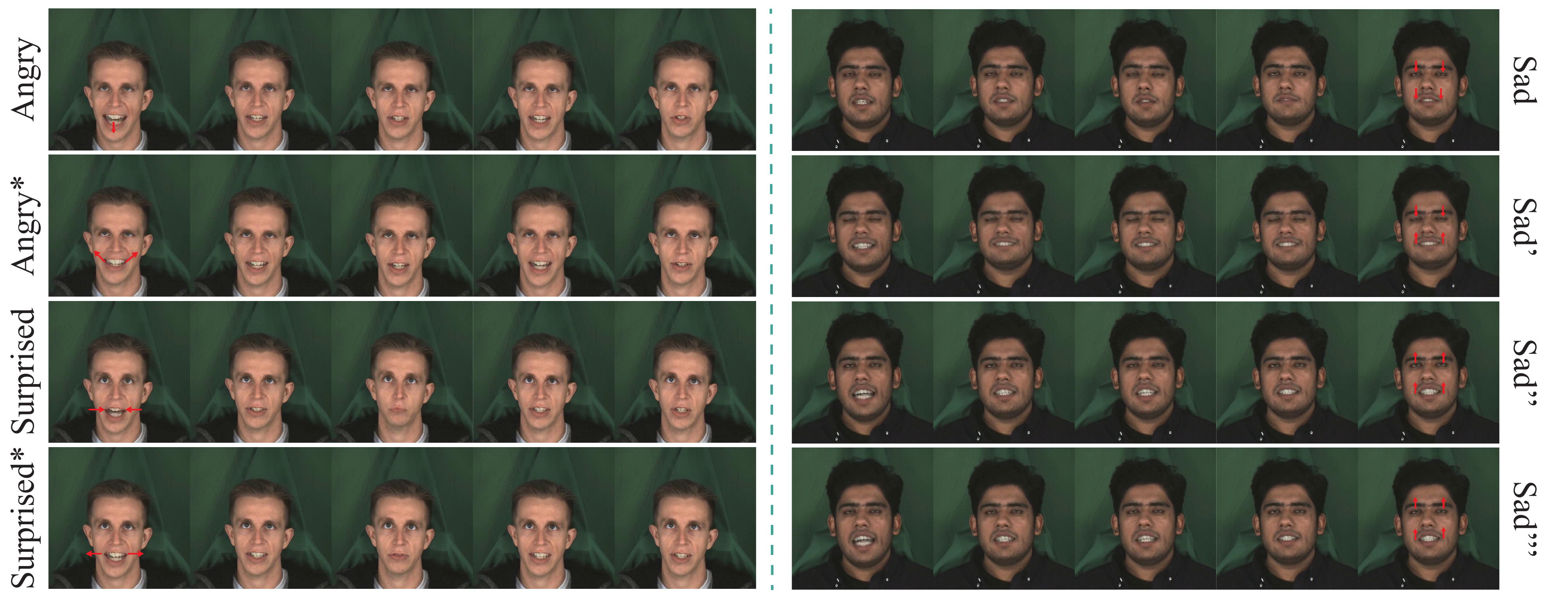} 
\caption{Generation in ablation study.}
\label{a-fig-ablation}
\end{figure*}
\begin{figure*}[h]
\centering
\includegraphics[width=\textwidth]{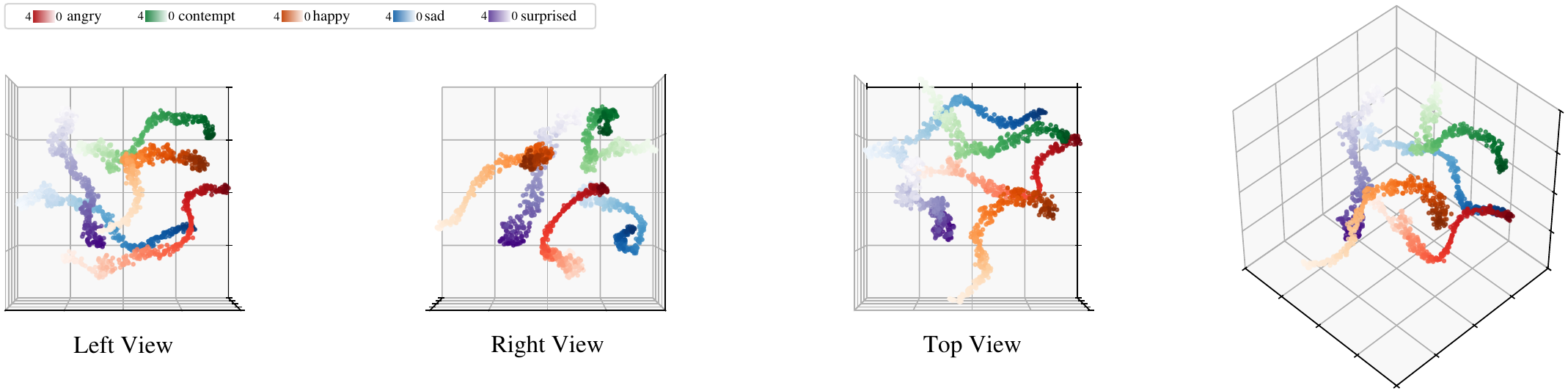} 
\caption{Multi-perspective visualization of the generated results using t-SNE downscaling.}
\label{fig:tsne}
\end{figure*}
As shown in Tab.~\ref{cdan-ab}, we utilized the Facial Landmark Distance (FLD) based on 3DMM coefficients, randomly selecting 200 frames for evaluation. Initially, we tested the \textit{Action Subspace} independently, using reference images with identical emotions and identities for generation. Subsequently, we incorporated the \textit{Isolation Subspace}, employing reference images with varying emotions and identities for generation. Replacing CDAN with a three-layer multilayer perceptron (MLP) or removing the serial connections resulted in a significant difference in FLD.

\subsection{Visual Emotion Effects}
Fig.~\ref{a-fig-ablation} illustrates the more detailed generative effects of LES-Talker compared to ablation results. Surprised* and angry* indicate the ablation where both levels of CDAN are replaced with a 3-layer MLP (consistent with Tab.\ref{cdan-ab}) . Sad', Sad'' and Sad''' respectively, indicate the absence of a serial connection between CDANs, the second level CDAN, and the special training strategy. Emotion level is set to 3.00. When either the CDANs, the special training strategy, or the serial connection is removed, the model loses the ability to express distinct emotional details, causing angry, surprised, and sad to resemble happy-like emotions, primarily by widening the mouth as the emotion level increases.
\subsection{Visualization of Emotion Representations}
We generated 1,755 videos across five emotions, with levels ranging from 0 to 4 in increments of 0.0114. Each video's frames were converted into 41-dimensional LES representations, averaged, and reduced using t-SNE. Fig.~\ref{fig:tsne} displays the multi-view spatial visualizations. For any pair of emotional point clouds, a separating plane exists in at least one view. Within each point cloud, intensity variations are clear and uniform. Our LES design achieves superior emotional isolation and fine-grained representation.

\section{Core Inference Code}
\subsection{Emotion Injector}
We present the inference process of the Emotion Injector, which primarily relies on the emotional semantics inherent in each vector within the designed LES space. The Emotion Injector utilizes the LES representations generated for each frame, along with 22 feature vectors stored in the \textit{Action Subspace} corresponding to eight emotions (including neutral), to perform emotion transformations. In the absence of an AU Source, a preset neutral emotion is employed as the initial LES representation.
\begin{algorithm}[htbp] 
\caption{Emo Injector Algorithm}
\label{alg:emo_injector}
\textbf{Input}: emo\_intensity, emo\_name, mode\_type, emo\_control\_pre\\
\textbf{Output}: emo\_control, point\_control
\begin{algorithmic}[1]
\STATE Load stats from CSV
\STATE Initialize emo\_control\_dict and point\_control\_dict
\IF{emo\_intensity is 0}
    \STATE emo\_control, point\_control = None, None
\ELSIF{emo\_intensity is -1}
    \STATE Use emo\_control\_pre or default AU values
\ELSIF{0 $<$ emo\_intensity $\leq$ 3}
    \STATE Calculate levels $i$ and $j$
    \STATE Set emo\_up\_name and emo\_down\_name
    \FOR{each AU}
        \STATE Interpolate emo\_control\_dict
        \IF{mode\_type is 3}
            \STATE Set point\_control\_dict
        \ENDIF
    \ENDFOR
\ELSE
    \STATE Calculate excess $ex$
    \FOR{each AU}
        \STATE Extrapolate emo\_control\_dict
        \IF{mode\_type is 3}
            \STATE Set point\_control\_dict
        \ENDIF
    \ENDFOR
\ENDIF
\end{algorithmic}
\end{algorithm}
\subsection{Inference Structure}
We present the inference algorithm, which operates in four modes: Mode 0 serves as the base; Mode 1 incorporates audio features to calibrate facial expressions; Mode 2 employs the full method by integrating complete multi-modal features as input; and Mode 3 operates without the AU Source. In Mode 3, to achieve more stable emotion transformations, we employ the point cloud obtained from the Emotion Injector as input to the \textit{Isolation Subspace}.

\begin{algorithm}[htbp] 
\caption{LES Core Structure}
\label{alg:les_core}
\textbf{Input}: current\_mel\_input, au, ref, emo\_type, emo\_intensity\\
\textbf{Output}: exp\_coeff\_pred
\begin{algorithmic}[1]
\IF{au is not None}
    \STATE Normalize au using means and stds
\ENDIF
\IF{mode\_type is 0}
    \STATE Compute exp\_coeff\_pred using CrossDimensionAttnNet1 with au and ref
\ELSIF{mode\_type is 1}
    \IF{emo\_control is not None}
        \STATE Adjust au with emo\_control
    \ENDIF
    \STATE Extract au45 from au
    \STATE Compute exp\_coeff\_pred using CrossDimensionAttnNet1 with au and ref
    \STATE Get pred\_ref from PretrainedNetG using current\_mel\_input, ref, au45
    \STATE Calculate exp\_resi using OffsetDecoder
    \STATE Add exp\_resi to exp\_coeff\_pred
\ELSIF{mode\_type is 2}
    \STATE Compute curr\_exp\_coeff\_pred using CrossDimensionAttnNet1 with au and ref
    \STATE Calculate emo\_resi using CrossDimensionAttnNet2
    \STATE Calculate exp\_resi and audio\_embed using OffsetDecoder
    \STATE Add emo\_resi And exp\_resi to curr\_exp\_coeff\_pred to get exp\_coeff\_pred
\ELSIF{mode\_type is 3}
    \STATE Extract au45 from au
    \STATE Get pred\_ref from PretrainedNetG using current\_mel\_input, ref, au45
    \STATE Calculate exp\_resi and audio\_embed using OffsetDecoder
    \STATE Predict aus using AuDecoder
    \STATE Adjust au45 and combine with aus
    \STATE Normalize aus
    \IF{emo\_control is not None}
        \STATE Adjust aus with emo\_control
    \ENDIF
    \STATE Compute curr\_exp\_coeff\_pred using CrossDimensionAttnNet1 with aus and ref
    \IF{point\_control is not None}
        \STATE Create au\_pred
        \STATE Calculate emo\_resi using CrossDimensionAttnNet2
        \STATE Add emo\_resi to curr\_exp\_coeff\_pred
    \ENDIF
    \STATE Add exp\_resi to curr\_exp\_coeff\_pred to get exp\_coeff\_pred
\ENDIF
\STATE \textbf{return} exp\_coeff\_pred
\end{algorithmic}
\end{algorithm}
\end{document}